\documentclass[letterpaper, 10 pt, conference]{sty/ieeeconf}
\usepackage{sty/macros}

\graphicspath{{figures/}}
\usepackage{cite}  % sorts/compresses citations: [1]--[3]
\makeatletter\let\NAT@parse\undefined\makeatother
\usepackage[hidelinks]{hyperref}  % links stay clickable, but no boxes or colors

\IEEEoverridecommandlockouts                              % This command is only needed if 
\title{\LARGE \bf
When Imperfect Automated Fact-Checking Backfires: Social Learning in the Presence of Misinformation
}

\author{Zixuan Liu and Ming Cao% <-this % stops a space
\thanks{This work was supported in part by the Dutch Research Council (NWO) through its Talent Programme (Vici-19902) and by the National Growth Fund through its AiNed (Artificial Intelligence Netherlands) programme (ELSA Lab for the Technical Industry, NGF.23.1608.005). The work of Liu was supported by the China Scholarship Council.}
\thanks{Zixuan Liu and Ming Cao are with the Faculty of Science and Engineering, University of Groningen, 9747 AG Groningen, The Netherlands
({\tt\small \{zixuan.liu, m.cao\}@rug.nl}).}% <-this % stops a space
}

\begin{document}

\maketitle
\thispagestyle{plain}
\pagestyle{plain}

%%%%%%%%%%%%%%%%%%%%%%%%%%%%%%%%%%%%%%%%%%%%%%%%%%%%%%%%%%%%%%%%%%%%%%%%%%%%%%%%

\begin{abstract}

Automated fact-checking tools, cheap and fast but imperfect, are increasingly available on social platforms. We study when such tools can backfire in a sequential social learning model of misinformation. Agents observe an article's reliability and their predecessors' actions, which indicate whether they liked or disliked the article. They may also pay to fact-check it. We examine how welfare, defined as the probability of a correct cascade, changes when a free, error-prone automated tool supplements costly but perfectly accurate manual fact-checking. Two opposing effects arise. Broader use of fact-checking helps agents escape wrong cascades. At the same time, errors from the automated tool can mislead agents who would otherwise have verified the article perfectly, reducing the likelihood of correct cascades. 
We prove that adding the tool strictly lowers welfare for articles of intermediate reliability when truthful articles are not too informative and manual checking is sufficiently costly. 
Simulations show that the tool can help on either side of the intermediate range, and that it improves on manual-only checking on average only above a threshold accuracy.

\end{abstract}

%%%%%%%%%%%%%%%%%%%%%%%%%%%%%%%%%%%%%%%%%%%%%%%%%%%%%%%%%%%%%%%%%%%%%%%%%%%%%%%%

\section{Introduction}

``Try AI companion'', ``ask anything''---social platforms increasingly invite users to talk to AI assistants built into their interfaces \cite{Yang24}, and many users already turn to them as fact-checking tools \cite{RML26, MWW26, MRJ26}. Such tools are quick and easy to use, but imperfect: their accuracy differs between true and false claims \cite{QuB24}, and they can even reduce users' ability to tell the two apart \cite{DeVerna24}. When a user consults a tool privately, others see only what she does, not what the tool told her, so a mistaken check can spread through actions and shape what a community comes to believe. This poses a design question for platforms: when does offering such a tool help collective learning, and when does it hurt?

% Social learning and cascades, then Pap20 and its limits, then the gap
This is a social learning question. Sequential social learning models have long been studied by economists \cite{Ban92, BHW92, SmS00, BHTW24}, and are also used to study online misinformation spread and fact-checking \cite{LWY26,Pap20}. In such models, typically, Bayesian agents sequentially make decisions based on their private signals and the observed actions of their predecessors. A key feature of these models is that, under appropriate conditions, an \emph{information cascade} occurs when agents follow others' actions and ignore their private information. In this case, not only does social learning stop, but agents could also herd on the wrong action, that is, a \emph{wrong cascade} can arise.

% What we do (manual vs automated, welfare, contrasts with Pap20)
In this paper, we study misinformation spread and fact-checking through a sequential social learning model. Agents arrive one after another and each likes or dislikes the same news article. They are uncertain about its veracity, which determines their payoff, but they know its reliability, the prior probability that it is truthful, and observe the actions of their predecessors. Before acting, an agent may fact-check the article. The check returns a private signal about the veracity, may be costly, and is subject to two-sided error rates. This differs from \cite{LWY26} and \cite{ChL26}, where fact-checking is public, and from \cite{Pap20}, where it is perfectly accurate. Our setting covers the situation where users fact-check privately, so a check informs others only through the action it leads to. 

% The reason to choose this setting is to better model the general paradigm shift of information acquisition, as people are more and more often using automated tools to assist decision making.   

We focus on two idealized fact-checking methods. \emph{Manual fact-checking} is a costly investigation that takes time and effort, but is error-free, as in \cite{Pap20}. \emph{Automated fact-checking}, such as asking an AI assistant, is free and easy to use but has two-sided error rates, so it can mislead. 
Idealizing manual fact-checking as error-free gives a clean baseline that isolates the effect of the automated tool's errors. 
We study how \emph{welfare}, defined as the probability of a correct cascade, changes when the automated tool is added alongside manual fact-checking, as when a platform builds it into its interface.

% contributions
The main contributions of the paper are as follows. 
We characterize when correct and wrong cascades arise with positive probability, via thresholds on the article's reliability that are explicit in the cost and error rates of the fact-checking method. 
We prove that adding the automated tool strictly decreases welfare for articles of reliability within an intermediate range, when manual fact-checking is sufficiently costly and truthful articles are not too informative, since the tool's errors can start wrong cascades that manual checking would have avoided.
Simulations show that welfare can also increase on both sides of this range if every agent who checks prefers the tool, and on one side if it replaces manual checking for only some agents. 
A further simulation varies the accuracy of the tool, showing that welfare falls once agents start to use it and exceeds the manual-only baseline only above a threshold accuracy, suggesting that a platform should not offer the tool below that accuracy.

\section{The Model} \label{sec:model}
% PLAN (OUTLINE.md): Sections II-A..II-E were tightened on 2026-10-07 to the
% 1.25-page model budget. The previous long version is in git (commit 870f146).

\subsection{State, News and Fact-Checking}
There is an underlying state of the world $\theta\in\{\stR,\stL\}$. Infinitely many agents, indexed by $t=1,2,\ldots$, arrive sequentially. Agent $t$ has a \emph{prior belief} $b_t$, the probability that $\theta=\stR$. Heterogeneity in $b_t$ captures differences in, e.g., ideology. The prior beliefs are independently drawn from a distribution with a cumulative distribution function $H$ that is continuous and strictly increasing on its support $[0,1]$.

A single news article, represented by a triple $(r,m,v)$, is generated at the beginning of time. It is first given a \emph{reliability} $r\in(0,1)$. Then its \emph{veracity} is drawn as $v=\vT$ (truthful) with probability $r$ or as $v=\vF$ (fake or fabricated) with probability $1-r$. A truthful article carries the \emph{message} $m=\theta$ with probability $p>1/2$, so it is informative about the state; a fake article carries $m=\stR$ with probability $q\in(0,1)$ whatever the state, so it is uninformative about the state. Agents can only observe $(r,m)$, but not $v$.\footnote{The reliability is often an observable feature of the article, such as its source or writing style.}  Variants of this news generating process are widely used in theoretical studies of misinformation spread~\cite{Pap20, MoS22, AOS24}. 

% with $r\in[0,1]$, $m\in\{\stR, \stL\}$ and $v\in\{\vT, \vF\}$ representing its \emph{reliability}, \emph{message} and \emph{veracity}, repectively. The article is first given a reliability $r$, then the veracity $v$ is drawn as $v=\vT$ with probability $r$ 

% which prescribes its \emph{veracity} $v$ being truthful, that is, the article is \emph{truthful} ($v=\vT$), with probability $r$; and \emph{fake} ($v=\vF$) with probability $1-r$. A truthful article carries the \emph{message} $m=\theta$ with probability $p>1/2$, so it is informative about the state; a fake article carries $m=\stR$ with probability $q\in(0,1)$ whatever the state, so it is uninformative. Similar settings are used in~\cite{Pap20, MoS22, AOS24}. Agents observe $(m,r)$, where $r$ stands for observable features of the article such as its source or writing style, but not $v$.

% PLAN [drop]: the news table; the paragraph above now states P[m | v, theta] in full.
% Commented out, unchanged.
% \begin{table}[tbp]
%     \centering
%     \tablestyle
%     \caption{The news distribution.} \label{tab:news}
%
%     \begin{tabular}{c|ccccc}
%     \hline
%     \multirow{2}{*}{$\P[m\given v,\theta]$}
%     & \multicolumn{2}{c}{Truthful $v=\vT$} && \multicolumn{2}{c}{Fake $v=\vF$} \\
%     \cline{2-3} \cline{5-6}
%     & $\theta=\stL$ & $\theta=\stR$ & & $\theta=\stL$ & $\theta=\stR$ \\
%     \cline{1-3} \cline{5-6}
%
%
%     $m=\stL$ & $p$   & $1-p$ & & $1-q$ & $1-q$ \\
%     $m=\stR$ & $1-p$ & $p$   & & $q$   & $q$ \\
%     \hline
%     \end{tabular}
%
% \end{table}

Agent $t$ takes an action $a_t\in\{\Like,\Dislike\}$, \emph{liking} or \emph{disliking} the article,\footnote{
    \emph{Like} and \emph{dislike} stand for positive and negative engagement in general, e.g., liking or sharing versus downvoting or commenting negatively. Unlike the \emph{sharing} action in~\cite{Pap20}, here an action does not affect who sees the article.
} with utility $u(\Like,v)=\1_{v=\vT}$ and $u(\Dislike,v)=\1_{v=\vF}$.
% \footnote{Unlike the model in \cite{Pap20}, actions here do not affect who observes the article.} 
Before acting, she may pay a cost $K\geq0$ to \emph{fact-check}, which returns a verdict $s_t\in\{\vT,\vF\}$ (the article is \emph{verified} or \emph{falsified}) with type-I and type-II error rates (following \cite{Shu17}, fake articles form the positive class):
% \footnote{
%     Following the fake news detection literature, a fake article is the positive class, so a type-I error is a false alarm and a type-II error is a miss \cite{Shu17}.
% }
\[
    e_1:=\P[s_t=\vF\given v=\vT],\quad e_2:=\P[s_t=\vT\given v=\vF],
\]
where $e_1,e_2\in[0,1)$ and $e_1+e_2<1$, so that $s_t$ is informative. Perfect fact-checking, $e_1=e_2=0$, is the case studied in~\cite{Pap20}. We treat fact-checking as an \emph{experiment} \cite{Ali18}: an agent checks when its value of information exceeds $K$ (see Section~\ref{subsec:individual}).\footnote{\cite{Ali18} assumes the outcome distributions under the two states ($v=\vT,\vF$ here) to be mutually absolutely continuous, which excludes $e_1=0$ or $e_2=0$. We thus take from it only the notions of an experiment and the value of information, not its results.
}

At time $t$, agent $t$ draws her prior belief $b_t$, observes the article $(r,m)$ and the action history $\calH_{t-1}:=(a_1,\ldots,a_{t-1})$ of her predecessors ($\calH_0=\emptyset$), decides whether to fact-check, observes $s_t$ privately if she does, and then acts. The distribution $H$, the parameters $p$, $q$, $K$, $e_1$, $e_2$, and the utility $u$ are common knowledge.

\subsubsection*{Discussion} Agents ``like'' (in the general sense) what they believe to be true and ``dislike'' what they believe to be fake, as in~\cite{Pap20,AOS24}. This is consistent with evidence that online engagement depends on perceived veracity~\cite{PeR21,PEM21}. The verdicts $s_1,s_2,\ldots$ are independent given the veracity, as in~\cite{Ali18}, reflecting the randomness in how each individual investigates, manually or with an automated tool. Correlated or even identical verdicts, such as the public signal of~\cite{LWY26}, are left to future work.

\subsection{The Individual Decision Problem} \label{subsec:individual}

Throughout the paper, we take the true state $\theta=\stR$, and in this section we illustrate with message $m=\stR$. The case $m=\stL$ can be treated analogously. Derivation details are in Appendix~\ref{app:details_model}.

\subsubsection*{Posterior Computation}
% Since the utility depends only on $v$, agents care about their posterior on veracity. 
Observing the article $(r, \stR)$, by Bayes' rule, agent $t$'s \emph{private belief} about its veracity is
\begin{align}
    \mu_t(b_t) :=~& \P[v=\vF\given m=\stR] \notag\\
    =~& \frac{(1-r) q}{(1-r) q + r\bigl( (1-p)(1-b_t) + p b_t \bigr)}. \label{eq:private_belief}
\end{align}
Since only a truthful article's message tracks the state, agent $t$'s assessment combines the reliability $r$ with her prior $b_t$: a message is more credible to an agent whose prior favors it. 

We focus on perfect Bayesian equilibrium, in which every agent knows all decision rules and can therefore compute the likelihood $\P_v[\calH_{t-1}]$ of any history, where $\P_v[\cdot]:=\P[\cdot\given m=\stR, v]$ is the probability conditional on message $\stR$ and veracity $v$.  
Let the \emph{public belief} $\rho_{t-1}:=\P_{\vF}[\calH_{t-1}]/\big(\P_{\vF}[\calH_{t-1}]+\P_{\vT}[\calH_{t-1}]\big)$ be the probability that the article is fake, given $\calH_{t-1}$ and a neutral prior on $v$. In particular, $\rho_0=1/2$ as $\calH_0=\emptyset$. Then, agent $t$'s \emph{posterior belief} that the article is fake is
\begin{align}
    \pi_t(\mu_t,\rho_{t-1}) :=~& \P[v=\vF\given \calH_{t-1}, m=\stR] \notag\\
    =~& \frac{\mu_t\rho_{t-1}}{\mu_t\rho_{t-1} + (1-\mu_t)(1-\rho_{t-1})}. 
    % \notag\\
    % &= \frac{(1-r)q\rho_{t-1}}{r\big((1-p)(1-b_t) + pb_t\big) (1-\rho_{t-1}) + (1-r)q\rho_{t-1}}. 
    \label{eq:posterior}
\end{align}

\subsubsection*{Actions and the Value of Fact-Checking}
\begin{figure}[tbp]
    \centering
    \vspace{4pt}% a float at the top of a page would otherwise start above the 0.75in margin
    % the PDF has the room of the y label on both sides, so that the plot frame is centred on the column
    \includegraphics{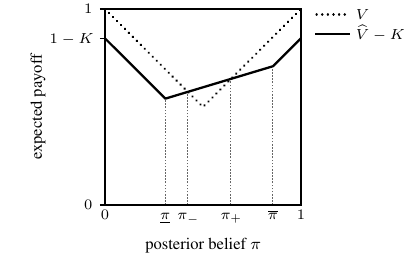}
    \caption{Expected payoff of an agent with posterior $\pi$ without fact-checking, ${V}$, and with it, $\widehat{V}-K$. Agents with $\pi\in(\underline{\pi},\overline{\pi})$ would follow the verdict, and those with $\pi\in(\pi_-,\pi_+)$, where $\widehat{V}-K>V$, also find it worth the cost and fact-check. Here $e_1=0.4$, $e_2=0.1$, $K=0.15$.}
    \label{fig:info_value}
\end{figure}

Without fact-checking, an agent with posterior $\pi$ maximizes her expected utility with $a^\star(\pi):=\arg\max_{a}\E_{v\sim\pi}[u(a,v)]$: she likes if $\pi\leq 0.5$ (ties go to liking, which have probability zero, as $H$ is continuous) and dislikes otherwise, which is worth ${V}(\pi):=\E_{v\sim\pi}[u(a^\star(\pi),v)]=\max\{\pi,1-\pi\}$. Let $\mathrm{BU}(\pi,s)$ be the Bayesian update of $\pi$ after the verdict $s$. The \emph{value of fact-checking} \cite{Ali18} is the expected utility of acting after observing $s$:
\begin{align}
    &\widehat{V}(\pi) := \E_{v\sim\pi}\Big[ \E_{s\given v}\big[u\big(a^\star(\mathrm{BU}(\pi, s)), v\big)\bigm| v\big] \Big] \notag\\
    &=\begin{cases}
        1-\pi & \text{if } \pi\leq \underline{\pi},\\
        \pi(1-e_2) + (1-\pi)(1-e_1)& \text{if } \underline{\pi} < \pi \leq \overline{\pi}, \\
        \pi & \text{if } \pi> \overline{\pi},
    \end{cases} \label{eq:V_pi}
\end{align}
where $\underline{\pi}=\frac{e_1}{1+e_1-e_2}$ and $\overline{\pi}=\frac{1-e_1}{1-e_1+e_2}$. An agent fact-checks if the gain is greater than the cost, i.e., $\widehat{V}(\pi)-{V}(\pi)> K$ (ties go to not checking). As Fig.~\ref{fig:info_value} shows, an agent with a near-certain posterior would not change her action whatever the verdict, hence has no motivation to check, whereas an ambivalent agent can possibly gain from checking by following the verdict. She does check if it is cheap enough, i.e., where $\widehat{V}-K$ lies above ${V}$.

\begin{proposition} \label{prop:cutoffs}
    If the cost $K<\frac{1}{2}(1-e_1-e_2)$, then there are posterior cutoffs $\pi_-=\frac{e_1+K}{1+e_1-e_2}$ and $\pi_+=\frac{1-e_1-K}{1-e_1+e_2}$ such that $\underline{\pi}\leq\pi_-<0.5<\pi_+\leq\overline{\pi}$, and an agent fact-checks if and only if her posterior $\pi\in(\pi_-,\pi_+)$. If conversely $K\geq \frac{1}{2}(1-e_1-e_2)$, then no agent fact-checks.
\end{proposition}
\begin{proof}
    Note that the gain of checking $\widehat{V}(\pi)-{V}(\pi)$ is maximized at $\pi=0.5$. Hence cutoffs $\pi_- <0.5< \pi_+$ with $\widehat{V}(\pi)-{V}(\pi)>K$ on $(\pi_-,\pi_+)$ exist if and only if $\widehat{V}(0.5)-{V}(0.5)>K$, i.e., $K<\frac{1}{2}(1-e_1-e_2)$. Solving $\widehat{V}(\pi) = {V}(\pi)+K$ gives the cutoff values.
\end{proof}

\subsubsection*{Cascade Sets of Public Beliefs}
Assume $K<\frac{1}{2}(1-e_1-e_2)$. With a fixed public belief $\rho$, the posterior $\pi(\mu(b),\rho)$ is strictly decreasing in the prior $b$, so the cutoffs of Proposition~\ref{prop:cutoffs} imply prior belief cutoffs $0\leq b_-(\rho)\leq b_+(\rho)\leq 1$: an agent dislikes if $b\in[0,b_-(\rho)]$, fact-checks and follows the result if $b\in(b_-(\rho),b_+(\rho))$, and likes if $b\in[b_+(\rho),1]$. Solving $\pi(\mu(b_\mp),\rho)=\pi_\pm$ for $b$ and truncating to $[0,1]$ gives the prior belief cutoffs:
% \begin{subequations}\label{eq:private_belief_cutoff_hi}
\medmuskip=1.5mu\relax % tighter binary-operator spacing so that (a) fits on its line
\begin{align}
    b_-(\rho) &= \bigg[ \frac{q(e_2+K)(1-r)\rho}{(2p-1)(1-e_1-K)r(1-\rho)}
    - \frac{1-p}{2p-1} \bigg]_{[0,1]},\allowdisplaybreaks \label{eq:private_belief_cutoff_lo}\\
    b_+(\rho) &= \bigg[ \frac{q(1-e_2-K)(1-r)\rho}{(2p-1)(e_1+K)r(1-\rho)}
    - \frac{1-p}{2p-1}\bigg]_{[0,1]}, \allowdisplaybreaks \label{eq:private_belief_cutoff_hi}
\end{align}
% \end{subequations}
% where $[\cdot]_{[0,1]}$ denotes the projection onto $[0,1]$. 
Both cutoffs are increasing in $\rho$, hence at extreme values, the whole support of $H$ is below $b_-(\rho)$ or above $b_+(\rho)$. We call the sets of such extreme public beliefs \emph{cascade sets}:
% \begin{subequations}\label{eq:public_belief_cascade}
\begin{align}
    J_\Like &:= \{\rho\given \mathrm{supp}(H)\subseteq[b_+(\rho),1]\} = \left[0, \rho_{b_+=0}^\stR\right], \label{eq:public_belief_cascade_L}\\
    J_\Dislike &:= \{\rho\given \mathrm{supp}(H)\subseteq[0, b_-(\rho)]\} = \left[\rho_{b_-=1}^\stR, 1\right], \label{eq:public_belief_cascade_D}
\end{align}
% \end{subequations}
where $\rho_{b_+=0}^\stR$ and $\rho_{b_-=1}^\stR$, solved from \eqref{eq:private_belief_cutoff_hi} and \eqref{eq:private_belief_cutoff_lo}, are public belief thresholds with respect to message $m=\stR$ (in odds, they are $\ell_\Like$ and $\ell_\Dislike$ in \eqref{eq:l_cutoffs_L}--\eqref{eq:l_cutoffs_D} below).

Intuitively, the map $b\mapsto \pi$ takes all prior beliefs to the liking posterior $[0,\pi_-]$ if and only if $\rho\in J_\Like$, and to $[\pi_+,1]$ if and only if $\rho\in J_\Dislike$. 
Once the public belief enters $J_\Like$ or $J_\Dislike$, all agents like or dislike, respectively, without checking. They ignore their private information and simply follow others, which is an \emph{information cascade} \cite{SmS00}. 
% The public belief may also only approach a cascade set; Remark~\ref{thm:cascade} treats this case.

A cascade set is a single point, $J_\Like=\{0\}$, if and only if $K=e_1=0$ (falsified verdict is free and conclusive),
% free and perfectly specific 
and $J_\Dislike=\{1\}$ if and only if $K=e_2=0$ (verified verdict is free and conclusive). 
% free and perfectly sensitive
% In the first case, a falsified verdict is free and conclusive, 
% since a truthful article is never falsified, so no agent likes the article without checking; 
% and the second case is symmetric. 
% This is close to the \emph{affordable overturning information} of~\cite{Ali18}. 
We rule out such ``magical'' methods, which cost nothing and are never wrong in at least one direction, but not the idealized methods of Section~\ref{sec:welfare}, each ideal in one respect only. Formally, we impose the following conditions on fact-checking methods.
\begin{assumption}\label{asmp:cost_error}
    The cost $K$ and the error rates $e_1,e_2$ satisfy: (a) $e_1+e_2<1$, (b) $K<\frac{1}{2}(1-e_1-e_2)$, and (c) $K>0$ or $e_1,e_2>0$.
\end{assumption}
Conditions (a) and (b) say that fact-checking is informative and worth its cost for some agents (Proposition~\ref{prop:cutoffs}), and (c) is exactly the exclusion above.

\subsection{Learning Dynamics}\label{subsec:learning_dynamics}

% Throughout this section, we impose the following assmptions on the cost and error rates of the fact-checking method.

% \begin{assumption}\label{asmp:cost_error}
% The cost $K\geq 0$ and error rates $e_1,e_2\in[0,1)$ of the fact-checking method satisfy the following conditions:
% \begin{enumerate}
%     \item The fact-checking method is informative: $e_1+e_2<1$.
%     \item The fact-checking is affordable: $K<(1-e_1-e_2)/2$.
%     \item The fact-checking is either costly or inaccurate in both directions: $K>0$ or $e_1, e_2>0$.
% \end{enumerate}
% \end{assumption}
% Under these assumptions, according to the previous analysis, the posterior cutoffs satisfy $0<\pi_-<0.5< \pi_+<1$. And the cascade sets of public beliefs $J_\Like$ and $J_\Dislike$ are non-singleton, i.e., $J_\Like=[0, \rho_{b_+=0}]$ and $J_\Dislike=[\rho_{b_-=1}, 1]$.

\subsubsection*{Markov Process of Likelihood Ratios} To facilitate the analysis, we describe the learning dynamics in terms of 
%  in what follows. Let $\rho_t$ be the public belief (on the article being fake) after agent $t$ chooses $a_t$ (initial public belief is neutral: $\rho_0=\frac{1}{2}$). We define 
% the \emph{public likelihood ratio} $\ell_t^{\vF}:=\frac{1-\rho_t}{\rho_t}=\frac{\P_{\vT}[\calH_t]}{\P_{\vF}[\calH_t]}$ (resp. $\ell_0^{\vT}:=1/\ell^{\vF}_t$), with initial value $\ell_0^{\vF}=\frac{1-\rho_0}{\rho_0}=1$ (resp. $\ell_0^{\vT}=1$). The stochastic process $\la\ell^{\vF}_t\ra_{t=0}^\infty$ (resp. $\la\ell_t^{\vT}\ra$) is a conditional supermartingale on $v=\vF$ (resp. $v=\vT$). We describe the dynamics of $\la\ell^{\vF}_t\ra$ by translating the previous notation into likelihood-ratio terms (the dynamics of $\la\ell^{\vT}_t\ra$ can be described similarly).
the \emph{public likelihood ratio} $\ell_t^{\vF}:=\frac{1-\rho_t}{\rho_t}=\frac{\P_{\vT}[\calH_t]}{\P_{\vF}[\calH_t]}$, with initial value $\ell_0^{\vF}=\frac{1-\rho_0}{\rho_0}=1$. Unless otherwise specified, we assume $v=\vF$ throughout the section, and illustrate the analysis of $(\ell^\vF_t)$. When $v=\vT$, the dynamics of $\ell^{\vT}_t:=1/\ell^{\vF}_t$ can be analyzed through symmetric arguments.

Denote the prior belief cutoffs by $\bar{b}_-(\ell^{\vF}_t):=b_-(\rho_t)$ and $\bar{b}_+(\ell^{\vF}_t):=b_+(\rho_t)$. Agent $t+1$ chooses $\Dislike$, to fact-check and follow the result, or $\Like$, whenever $b_{t+1}$ lies in $[0, \bb_-(\ell^{\vF}_t)]$, $(\bar{b}_-(\ell^{\vF}_t),\bar{b}_+(\ell^{\vF}_t))$, or $[\bar{b}_+(\ell^{\vF}_t),1]$, respectively. After her action $a_{t+1}$, the public likelihood ratio will be updated to $
\ell^{\vF}_{t+1} = \frac{\P_{\vT}[\calH_{t}\cup \{a_{t+1}\}]}{\P_{\vF}[\calH_{t}\cup \{a_{t+1}\}]} 
= \ell^{\vF}_{t}\cdot \frac{\P_{\vT}[a_{t+1}\given \calH_t]}{\P_{\vF}[a_{t+1}\given \calH_t]}$,\footnote{
    Note that when $v=\vF$, the denominator $\P_{\vF}[a_{t+1}\given \calH_t]$ is nonzero, since otherwise agent $t+1$ can never choose $a_{t+1}$. So the update is well-defined. The same applies to the analysis of $\ell^\vT_t$ when $v=\vT$. 
} that is,
\begin{align}
    \ell^{\vF}_{t+1} =~& \begin{cases}
        \ell^{\vF}_t \cdot \frac{1-G_{\vT}(\ell^{\vF}_t)}{1-G_{\vF}(\ell^{\vF}_t)}, &\text{if }a_{t+1}=\Like, \\
        \ell^{\vF}_t \cdot \frac{G_{\vT}(\ell^{\vF}_t)}{G_{\vF}(\ell^{\vF}_t)}, &\text{if }a_{t+1}=\Dislike, 
    \end{cases}\label{eq:l_dynamics_SN}
\end{align}
where, for $v\in\{\vT,\vF\}$, $G_v(\ell^{\vF}_t):=\P_v[a_{t+1}=\Dislike\given\calH_t]=H(\bb_-(\ell^{\vF}_t))+c_v\Phi(\ell^{\vF}_t)$, with $c_\vT:=e_1$, $c_\vF:=1-e_2$ (the pro\-bability that a check yields $\Dislike$) and $\Phi(\ell):=H(\bb_+)-H(\bb_-)$ the probability of fact-checking.
Therefore, $(a_t,\, \ell^{\vF}_t )$ is a discrete-time Markov process on $\{\Like,\Dislike\}\times[0,\infty)$: it moves to $(\Like,\ell^{\vF}_{t+1})$ with probability $1-G_\vF(\ell^{\vF}_t)$ and to $(\Dislike,\ell^{\vF}_{t+1})$ with probability $G_\vF(\ell^{\vF}_t)$, where $\ell^{\vF}_{t+1}$ is given by \eqref{eq:l_dynamics_SN}.
% \begin{align*}
%     \la a_t, \ell^{\vF}_t\ra &\rightarrow \la \Like, \ell^{\vF}_{t+1}\ra 
%     &&\text{with prob. } 1-G_{\vT}(\ell^{\vF}_t) &&\text{ if } v=\vT;
%     \quad 1-G_{\vF}(\ell^{\vF}_t) &&\text{ if } v=\vF,\\
%     \la a_t, \ell^{\vF}_t\ra &\rightarrow \la \Dislike, \ell^{\vF}_{t+1}\ra 
%     &&\text{with prob. } G_{\vT}(\ell^{\vF}_t) &&\text{ if } v=\vT;
%     \quad G_{\vF}(\ell^{\vF}_t) &&\text{ if } v=\vF.
% \end{align*}
% \begin{align*}
%     \text{If } v=\vT, && \la a_t, \ell^{\vF}_t\ra &\rightarrow \begin{cases}
%         \la \Like, \ell^{\vF}_{t+1}\ra &\text{with prob. } 1-G_{\vT}(\ell^{\vF}_t), \\
%         \la \Dislike, \ell^{\vF}_{t+1}\ra &\text{with prob. } G_{\vT}(\ell^{\vF}_t);
%     \end{cases}\\
%     \text{if } v=\vF, && \la a_t, \ell^{\vF}_t\ra &\rightarrow \begin{cases}
%         \la \Like, \ell^{\vF}_{t+1}\ra &\text{with prob. } 1-G_{\vF}(\ell^{\vF}_t), \\
%         \la \Dislike, \ell^{\vF}_{t+1}\ra &\text{with prob. } G_{\vF}(\ell^{\vF}_t).
%     \end{cases}
% \end{align*}

\subsubsection*{Information Cascades} To rephrase the cascade sets in terms of public likelihood ratios, define
% \begin{subequations}\label{eq:l_cutoffs}
\begin{align}
    \ell_\Like&:=\frac{1-\rho^\stR_{b_+=0}}{\rho^\stR_{b_+=0}}=\frac{q(1-r)(1-e_2-K)}{(1-p) r (e_1+K)}, \label{eq:l_cutoffs_L}\\
    \ell_\Dislike&:=\frac{1-\rho^\stR_{b_-=1}}{\rho^\stR_{b_-=1}}=\frac{q(1-r)(e_2+K)  }{p     r (1-e_1-K)}. \label{eq:l_cutoffs_D}
\end{align}
% \end{subequations}
By \eqref{eq:public_belief_cascade_L} and \eqref{eq:public_belief_cascade_D}, the cascade sets with respect to $\ell_t^\vF$ are $\bar{J}_\Like := [\ell_\Like, \infty ]$ and $\bar{J}_\Dislike := [0, \ell_\Dislike]$. These sets are absorbing: by \eqref{eq:l_dynamics_SN}, $\ell^\vF_t$ freezes once it enters them, so a cascade arises at the start if $\ell^\vF_0=1\in\bar J_\Dislike\cup\bar J_\Like$. Otherwise, $\ell^\vF_t$ converges almost surely (a.s.), i.e., with $\P_\vF$-probability one, and its limit $\ell^\vF_\infty$ lies in $\bar J_\Dislike\cup\bar J_\Like\setminus \{\infty\}$.\footnote{
This includes the case where $\ell^\vF_t$ only approaches a cascade set asymptotically: agents then still fact-check with vanishing probability, but eventually all take the same action \cite{SmS00}.
The proofs are given in Lemmas~\ref{lem:mart}--\ref{lem:stationary}.}
\begin{definition}[cascade]\label{def:cascade}
    A $\Dislike$ \emph{cascade} arises if $\ell^\vF_\infty\in\bar J_\Dislike$, and an $\Like$ \emph{cascade} arises if $\ell^\vF_\infty\in\bar J_\Like$. A cascade is \emph{correct} if it matches the veracity, i.e., an $\Like$ cascade if $v=\vT$, a $\Dislike$ cascade if $v=\vF$.
\end{definition} 
% since near a cascade set an opposite action can only come from a fact-check and moves $\ell^\vF_t$ by a fixed factor.
% given $v=\vF$, $(\ell^{\vF}_t)$ is a nonnegative supermartingale (Lemma~\ref{lem:mart}),\footnote{\color{red}
%     TODO: I still need to explain this in more detail, since most of social learning literature use it as a conditional maritingale. However, in our model, the fact-checking signal distributions are not necessarily mutually absolutely continuous, the Radon-Nikodym derivative does not exist everywhere.
% } hence it converges to $\ell^{\vF}_\infty\in[0, \infty)$ almost surely. Moreover, it can only converge to $\bar{J}_\Dislike\cup\bar{J}_\Like\setminus\{\infty\}$ (Lemma~\ref{lem:stationary}).

In the following, we characterize the learning outcomes when $(\ell^\vF_t)$ starts from the non-cascade interval.
% To do so, we need two more thresholds inside $(\ell_\Dislike, \ell_\Like)$:
% \begin{subequations}\label{eq:l_cutoffs_MN}
% \begin{align}
%     M &:= \frac{1-\rho^\stR_{b_+=1}}{\rho^\stR_{b_+=1}} = \frac{q(1-r)(1-e_2-K)}{p\,r\,(e_1+K)},\\
%     N &:= \frac{1-\rho^\stR_{b_-=0}}{\rho^\stR_{b_-=0}} = \frac{q(1-r)(e_2+K)}{(1-p)\,r\,(1-e_1-K)},
% \end{align}
% \end{subequations}
% where thresholds $\rho^\stR_{b_+=1}$ and $\rho^\stR_{b_-=0}$ are given in Appendix~\ref{app:public_belief_thresholds}. Since $\bar{b}_-$ and $\bar{b}_+$ are decreasing in $\ell$, we have $\bar{b}_+(\ell)=1\iff\ell\le M$ and $\bar{b}_-(\ell)=0\iff\ell\ge N$. Therefore, on $(\ell_\Dislike,M]$,
% nobody likes without checking: agents either choose $\Dislike$ directly or fact-check and follow the result. Similarly, on $[N,\ell_\Like)$, nobody dislikes without checking. If moreover $e_2=0$, a fake article never passes a fact-check, so under $v=\vF$ nobody likes on $(\ell_\Dislike,M]$ and $\ell^\vF_t$ can only fall toward $\bar J_\Dislike$. This drives Theorem~\ref{thm:learning_outcomes}(a), and (b) is symmetric.

\begin{theorem}\label{thm:learning_outcomes}
    Suppose Assumption~\ref{asmp:cost_error} holds and $\ell^{\vF}_0\in(\ell_\Dislike, \ell_\Like)$. 
    \begin{enumerate}[label=(\alph*)]
        \item Suppose that $v=\vF$ and $e_2=0$. If $\ell^{\vF}_0\leq M$, then a $\Dislike$ cascade arises a.s.; otherwise, both $\Like$ and $\Dislike$ cascades arise with positive probability.
        \item Suppose that $v=\vT$ and $e_1=0$. If $\ell^{\vF}_0\geq N$, then an $\Like$ cascade arises a.s.; otherwise, both $\Like$ and $\Dislike$ cascades arise with positive probability.
        \item In all other cases, both $\Like$ and $\Dislike$ cascades arise with positive probability.
    \end{enumerate}
    Here, $M := \frac{q(1-r)(1-e_2-K)}{p\,r\,(e_1+K)}$ and 
    $N := \frac{q(1-r)(e_2+K)}{(1-p)\,r\,(1-e_1-K)}$.
\end{theorem}
\begin{proof}[Proof idea]
    Since $\bar{b}_-$ and $\bar{b}_+$ are decreasing in $\ell$, we have $\bar{b}_+(\ell)=1\iff\ell\le M$; $\bar{b}_-(\ell)=0\iff\ell\ge\nobreak N$. Therefore, on $(\ell_\Dislike,M]$,
    nobody likes without checking, and on $[N,\ell_\Like)$, nobody dislikes without checking. If moreover $e_2=0$, a fake article never passes a fact-check, so under $v=\vF$ nobody likes on $(\ell_\Dislike,M]$ and $\ell^\vF_t$ can only fall toward $\bar J_\Dislike$. This drives (a), and (b) is symmetric. The positive probability arguments follow from the (super)martingale property of $(\ell_t^{\vF})$. See Appendix~\ref{app:proof_learning_outcomes} for the complete proof.
\end{proof}

% With the characterization of learning outcomes in terms of the likelihood ratio thresholds, it is a short step to relate the reliability $r$ to learning outcomes. To do so, define the \emph{unconditional reliability thresholds} of a fact-checking method $\calM:=(K,e_1,e_2)$ as
% % \begin{subequations}\label{eq:r}
% \begin{align}
%     r_\Dislike(\calM)&:=\frac{q(e_2+K)}{p(1-e_1-K)+q(e_2+K)},\label{eq:r}\\
%     r_\Like(\calM)&:=\frac{q(1-e_2-K)}{(1-p)(e_1+K)+q(1-e_2-K)},
% \end{align}
% % \end{subequations}
% and the \emph{conditional reliability thresholds} as: 
% % \begin{subequations}\label{eq:r_cond}
% \begin{align}
%     \widehat{r}_\Dislike(\calM)&:=\frac{q(1-e_2-K)}{p(e_1+K)+q(1-e_2-K)},\\
%     \widehat{r}_\Like(\calM)&:=\frac{q(e_2+K)}{(1-p)(1-e_1-K)+q(e_2+K)}.\label{eq:r_cond}
% \end{align}
% % \end{subequations}

Theorem~\ref{thm:learning_outcomes} readily translates into thresholds on the reliability $r$. Denote a fact-checking method by $\calM:=(K,e_1,e_2)$. Since the thresholds $\ell_\Dislike,\ell_\Like,M,N$ are strictly decreasing in $r$, each equals $\ell^\vF_0=1$ at exactly one reliability. We denote these by $r_\Dislike(\calM)$, $r_\Like(\calM)$, $\widehat r_\Dislike(\calM)$, $\widehat r_\Like(\calM)$, respectively, given as:
\begin{align}
    r_\Dislike(\calM) &=\frac{q(e_2+K)}{p(1-e_1-K)+q(e_2+K)},\label{eq:r}\\
    r_\Like(\calM) &=\frac{q(1-e_2-K)}{(1-p)(e_1+K)+q(1-e_2-K)},\\
    \widehat{r}_\Dislike(\calM) &=\frac{q(1-e_2-K)}{p(e_1+K)+q(1-e_2-K)},\label{eq:r_cond_D}\\
    \widehat{r}_\Like(\calM) &=\frac{q(e_2+K)}{(1-p)(1-e_1-K)+q(e_2+K)}.\label{eq:r_cond}
\end{align}

\begin{corollary}\label{cor:reliability_thresholds}
    Suppose Assumption~\ref{asmp:cost_error} holds. There exist reliability thresholds $0<\underline{r}<\overline{r}<1$ depending on $v$ and $\calM$, such that, conditional on $v$,
    \begin{enumerate}[label=(\alph*)]
        \item If $r\geq\overline{r}$ (highly reliable), an $\Like$ cascade arises a.s.
        \item If $r\leq\underline{r}$ (poorly reliable), a $\Dislike$ cascade arises a.s.
        \item If $\underline{r}<r<\overline{r}$ (moderately reliable), both $\Like$ and $\Dislike$ cascades arise with positive probability.
    \end{enumerate}
    Specifically, the thresholds $(\underline{r},\, \overline{r})$ are $(\widehat{r}_\Dislike,\, r_\Like)$ if $v=\vF, e_2=0$;
    $({r}_\Dislike,\, \widehat{r}_\Like)$ if $v=\vT, e_1=0$; and 
    $({r}_\Dislike,\, r_\Like)$ otherwise.
    % \begin{align*}
    %     (\underline{r},\, \overline{r})&= \begin{cases}
    %         \big(\widehat{r}_\Dislike(\calM),\, r_\Like(\calM)\big), & \text{ if } v=\vF \text{ and } e_2=0,\\
    %         \big({r}_\Dislike(\calM),\, \widehat{r}_\Like(\calM)\big), & \text{ if } v=\vT \text{ and } e_1=0,\\
    %         \big({r}_\Dislike(\calM),\, r_\Like(\calM)\big), & \text{ otherwise.}
    %     \end{cases}
    % \end{align*}
\end{corollary}
% \begin{proof}
%     Notice that the likelihood ratio thresholds $\ell_\Like,\ell_\Dislike, M$ and $N$ are all strictly decreasing with $r$. Then, by Theorem~\ref{thm:learning_outcomes} and $\ell_0^\vF=1$, we have the result.
% \end{proof}

\begin{proof}
    For $r\le r_\Dislike$ (resp.\ $r\ge r_\Like$), even an agent with $b=1$ (resp.\ $b=0$) dislikes (likes) without checking at $t=1$, so a $\Dislike$ ($\Like$) cascade starts at once, whatever $v$ is. For $r\le\widehat r_\Dislike$ (resp.\ $r\ge\widehat r_\Like$), nobody likes (dislikes) without checking at $t=1$. If moreover $v=\vF$ and $e_2=0$ (resp.\ $v=\vT$ and $e_1=0$), a check never contradicts a $\Dislike$ ($\Like$) cascade, so it still arises a.s. 
    Since $M,N\in(\ell_\Dislike,\ell_\Like)$ by Assumption~\ref{asmp:cost_error}(b) and $p>\frac12$, we have $\widehat r_\Dislike,\widehat r_\Like\in(r_\Dislike,r_\Like)$, and Theorem~\ref{thm:learning_outcomes} with $\ell^\vF_0=1$ gives the result.
\end{proof}

% The corollary has an implication: with a cheaper fact-checking tool, the reliability interval $(\underline{r},\overline{r})$ is larger, i.e., a wider range of moderately reliable articles do not cause an immediate cascade at $t=1$, regardless of whether the article is truthful or fake.

\begin{remark}
    In the social learning literature, \emph{learning is complete} if $\ell^{\vF}_\infty = 0$ a.s. when $v=\vF$ and $\ell^{\vT}_\infty = 0$ a.s. when $v=\vT$ \cite{SmS00}. In our model, this holds if and only if fact-checking is free \emph{and} perfectly accurate in both directions. But this is excluded by Assumption~\ref{asmp:cost_error}, hence \emph{learning is incomplete}.
\end{remark}

% PLAN [drop]: open questions for the journal version (the paper now proves results
% instead of relying mainly on simulation). Commented out, unchanged.
% In the latter (more realistic) case, it is interesting to ask:
% \begin{enumerate}
%     \item If no cascade starts at $t=1$, how long does it take for one to start? Do truthful / fake articles take different times to form good / bad cascades?
%     \item If both good and bad cascades arise with positive probability, how does the probabilities change with error rates $e_1, e_2$ and cost $K$?
% \end{enumerate}
%
%
% These questions do not appear tractable analytically.\footnote{
%     \cite{LSB16} analyze some cases of a similar sequential model. However, in their model there is no continuous private belief, only a binary private signal, so the Markov process has a ``counting structure'' that makes the analysis manageable in some cases. In our model, the continuous private belief makes the transition probabilities (outside the cascade sets) depend on $\ell_t$, unlike the constant probabilities in theirs.
% } A full treatment appears to require tools for Markov processes on infinite state spaces, in particular for computing limit distributions. For now, we rely mainly on Monte Carlo simulation to study how parameter changes affect the learning outcome.

\section{Problem of Welfare Change}\label{sec:welfare}
\subsection{Welfare Definition}

The learning outcome is evaluated by \emph{welfare}, the probability of a correct cascade. A finite set $\bbM$ of fact-checking methods is available. When $\bbM$ has several methods, an agent fact-checks with at most one of them: the $\calM$ maximizing $\widehat V_\calM(\pi)-K_\calM$, or none if no $\calM$ beats $V(\pi)$.

\begin{definition}[welfare]\label{def:welfare}
    Let $\calE:=(p,q,r,H)$ be the information environment, $\bbM$ be the set of available methods, and $C$ be the event that a correct cascade arises. The \emph{welfare} of $\calE$ with methods $\bbM$ is
    % \begin{align*}
        $W(\calE,\bbM):=\P[C]$.
    % \end{align*}
    For a realized article $(\bar m,\bar v)\in\{\stL,\stR\}\times\{\vT,\vF\}$, its \emph{welfare} is $\lam^{\bar m,\bar v}(\calE,\bbM):=\P[C\given m=\bar m,v=\bar v]$.
\end{definition}
Note that $W(\calE,\bbM)=\E[\lam^{m,v}(\calE,\bbM)]$. Since $\lam^{m,v}$ is a random variable that depends on $(m,v)$, by the law of total probability, we have 
\begin{multline*}
W(\calE,\bbM) = rp\cdot\lam^{\stR,\vT} + r(1-p)\cdot\lam^{\stL,\vT} \\
 + (1-r)q\cdot\lam^{\stR,\vF} + (1-r)(1-q)\cdot\lam^{\stL,\vF}.
\end{multline*}

\begin{assumption}\label{asmp:symmetric_news}
    The prior belief distribution $H$ is symmetric: $1-H(x)=H(1-x)$, and the shares of $\stL$ and $\stR$ articles in fake news are equal: $q=0.5$.
\end{assumption}
Under this assumption, it is straightforward to see the symmetry between the dynamics of $\stL$ and $\stR$ articles, so $\lam^{\stR,\bar{v}}=\lam^{\stL,\bar{v}}\ \forall \bar{v}\in\{\vT,\vF\}$. We therefore drop the superscript $\stR$ and write $\lam^{v}:=\lam^{\stR,v}$ for $v\in\{\vT,\vF\}$. Then, the welfare decomposition can be further simplified as:
\begin{gather}
W(\calE,\bbM)
= r\cdot \lam^{\vT} + (1-r) \cdot \lam^{\vF}. \label{eq:symmetric_welfare}
\end{gather} 
We restrict the analysis to this symmetric case, as it greatly simplifies notation. The asymmetric case can be analyzed in like manner, where thresholds and magnitudes change, but the qualitative results do not.

\subsection{Problem Setup}

We focus on two types of idealized fact-checking methods, both satisfying Assumption~\ref{asmp:cost_error}.

The first method is the costly, perfectly accurate \emph{manual fact-checking} $\MH=(\KH,0,0)$ with $\KH\in(0,0.5)$, so it is affordable for some agents (Proposition~\ref{prop:cutoffs}). 
It is costly because an investigation generally takes time and effort. The perfect accuracy is an idealized benchmark.\footnote{
We expect the qualitative findings of Section~\ref{sec:MC} to persist when $\MH$ has small error rates, but our theoretical results do not cover this case. 
}

The second method is the free, error-prone \emph{automated fact-checking} $\MA=(\KA,\eA{1},\eA{2})$ with $\KA=0$, $\eA{1},\eA{2}\in(0,1)$ and $\eA{1}+\eA{2}< 1$. We also assume that $\eA{1}>\eA{2}$,\footnote{
    $\eA{1}>\eA{2}$ also excludes the case that $\KH=\eA{1}=\eA{2}$. Thus our tie-breaking rule between methods (agents prefer $\MH$ to $\MA$ when two methods have the same expected payoff) only needs to work on a population of measure zero. Hence the results are robust to other rules.
}  since automated checkers identify false claims more reliably than true ones~\cite{QuB24,DeVerna24}.

We study the problem: 
% if an additional automated fact-checking method is introduced alongside the existing manual method, how will welfare change?
% Suppose that an additional automated fact-checking method $\MA$ is introduced alongside the existing manual fact-checking method $\MH$, how will welfare change? 
\begin{center}
\emph{If an additional automated fact-checking method\\ is introduced alongside the existing manual method, \\how will welfare change?}
\end{center}
Formally, we seek to characterize the welfare change:
\begin{align}
    &\Delta W := W(\calE,\{\MH,\MA\}) - W(\calE,\{\MH\}), \label{eq:welfare_change}
\end{align}
and to understand how it depends on the parameters
$p$, $r$, $H$, $\KH$, $\eA{1}$, and $\eA{2}$. Note that, under Assumption~\ref{asmp:symmetric_news}, by \eqref{eq:symmetric_welfare}, the welfare change can be written as:
% \begin{align*}
\(
        \Delta W = r \Delta W_\vT + (1-r) \Delta W_\vF,
\)
% \end{align*}
% r\cdot \lam^{\vT} + (1-r) \cdot \lam^{\vF}
where $\Delta W_\vT :=\lam^{\vT}(\{\MH,\MA\})-\lam^{\vT}(\{\MH\})$ and $\Delta W_\vF := \lam^{\vF}(\{\MH,\MA\})-\lam^{\vF}(\{\MH\})$.
    % where $\Delta W_\vT := \Delta w(\stR,\vT)=2\Delta \lam_\Like^{\stR,\vT}$ and $\Delta W_\vF := \linebreak\Delta w(\stR,\vF)= -2\Delta\lam_\Like^{\stR,\vF}$, and $\Delta\lam_\Like^{m,v}:=\lam_\Like^{m,v}(\{\MH,\MA\})\linebreak - \lam_\Like^{m,v}(\MH)$ denotes the change of probability that an article $(m,v)$ triggers an $\Like$ cascade.

% \begin{align*}
%     \Delta W^{m,\vT} &= 2\Delta \lam_\Like^{m,\vT},\\
%     \Delta W^{m,\vF} &= -2\Delta\lam_\Like^{m,\vF},\qquad m = \stR,\stL,
% \end{align*}
% where $\Delta\lam_\Like^{m,v}:=\lam_\Like^{m,v}(\{\MH,\MA\}) - \lam_\Like^{m,v}(\MH)$ denotes the change of probability that an article $(m,v)$ triggers an $\Like$ cascade (good for the welfare of truthful news, bad for that of fake news) when $\MA$ is additionally introduced alongside $\MH$.

\section{Simulations and Analysis} \label{sec:MC}

We study the welfare change \eqref{eq:welfare_change} by Monte Carlo simulation and theoretically certify its sign in some parameter regions. Depending on the accuracy of $\MA$ and the cost of $\MH$, there are three regimes, according to who uses $\MA$:
\begin{enumerate}[label=(\roman*)]
    \item\label{it:manual_preferred} $\KH \leq \frac{\eA{2}}{1-\eA{1}+\eA{2}}$: no one uses $\MA$, so $\Delta W = 0$.
    \item\label{it:both_used} $\frac{\eA{2}}{1-\eA{1}+\eA{2}} < \KH \leq \frac{\eA{1}}{1+\eA{1}-\eA{2}}$: both methods are used, depending on the posterior (Fig.~\ref{fig:sec5_cutoffs_moderate_tech}).
    \item\label{it:automated_preferred} $\KH > \frac{\eA{1}}{1+\eA{1}-\eA{2}}$: every agent who checks prefers $\MA$ (Fig.~\ref{fig:sec5_cutoffs_strong_tech}).
\end{enumerate}
Section~\ref{subsec:welfare_case3} treats regime~\ref{it:automated_preferred}, the main regime, and Section~\ref{subsubsec:asym_acc_two_tools} regime~\ref{it:both_used}. Section~\ref{subsec:improving_tech} explores the welfare change as the accuracy of $\MA$ improves.

% ------------------------- MT more favorable --------------------------------------
\subsection{Automated Method Preferred}\label{subsec:welfare_case3}

In regime~\ref{it:automated_preferred}, only $\MA$ is used, so the dynamics are those of Section~\ref{sec:model} for the single method $\MA$, and $\Delta W = W(\calE, \{\MA\})-W(\calE, \{\MH\})$.

\subsubsection*{Simulation Setup}
\begin{figure}[tbp]
\centering
\vspace{4pt}% a float at the top of a page would otherwise start above the 0.75in margin
% centre the subcaptions on the plot area: the margins are _L, _R of simulation/fig3_welfare_r_p.py
\captionsetup[sub]{margin={0.34in,0.04in}}
\begin{subfigure}[t]{0.49\columnwidth}
\includegraphics[width=\linewidth]{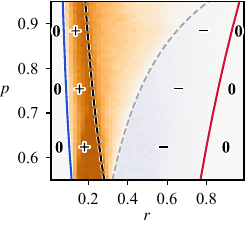}
\caption{$\Delta W_\vT$}\label{fig:sec5_d_wel_rho0_p_T}
\end{subfigure}\hfill
\begin{subfigure}[t]{0.49\columnwidth}
\includegraphics[width=\linewidth]{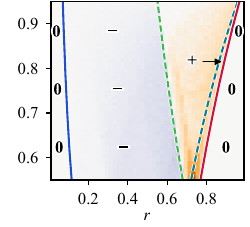}
\caption{$\Delta W_\vF$}\label{fig:sec5_d_wel_rho0_p_F}
\end{subfigure}\\[2pt]
\begin{subfigure}[t]{0.49\columnwidth}
\includegraphics[width=\linewidth]{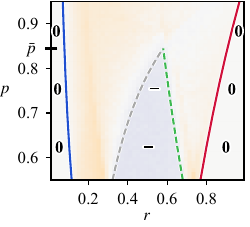}
\caption{$\Delta W$}\label{fig:sec5_d_wel_rho0_p_all}
\end{subfigure}\hfill
\begin{subfigure}[t]{0.49\columnwidth}
\includegraphics[width=\linewidth]{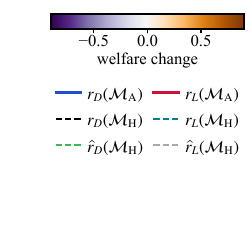}
\end{subfigure}
\caption{Welfare change in regime~\ref{it:automated_preferred}: $\KH=0.3$, $\eA{1}=0.3$, $\eA{2}=0.1$. Lines: the thresholds \eqref{eq:r}--\eqref{eq:r_cond} bounding the regions where Proposition~\ref{prop:sec4_shaded_area} certifies the sign ($+$, $-$), and $0$ where a cascade starts at $t=1$ under both method sets. Bold tick marks $\bar p$.}\label{fig:sec5_d_wel_rho0_p}
\end{figure}

Let $\eA{1}=0.3$, $\eA{2}=0.1$, $\KH=0.3$, and $H$ be a beta distribution with parameters $0.5, 0.5$. On a $50\times 50$ grid of $r\in[0.01,0.99]$ and $p\in[0.55,0.95]$, we estimate the cascade probabilities by their frequencies in $2000$ runs at each grid point (Fig.~\ref{fig:sec5_d_wel_rho0_p}). 
A run ends as a cascade once the next agent's probability of liking, or of disliking, is below $10^{-5}$ under the veracity of the article. By this numerical criterion, all runs cascade within $t=300$.
% The frequencies agree with a numerical solution of the Markov chain to within sampling error (mean difference below $10^{-3}$). 
The following proposition certifies the sign of $\Delta W$ in some regions of the $(r,p)$-plane.

\begin{proposition} \label{prop:sec4_shaded_area}
    Suppose that Assumption~\ref{asmp:symmetric_news} holds and $\KH > \frac{\eA{1}}{1+\eA{1}-\eA{2}}$. We have:
    \begin{enumerate}[label=(\alph*)]
    \item $\Delta W_\vT=\Delta W_\vF=0$ for $r\notin\big(r_\Dislike(\MA),r_\Like(\MA)\big)$.
    
    \item $\Delta W_{\vT}>0$ on $\big(r_\Dislike(\MA),\,r_\Dislike(\MH)\big)$; 
    $\Delta W_{\vT}<0$ on $\big(\widehat{r}_\Like(\MH),\,r_\Like(\MA)\big)$.
    
    \item $\Delta W_{\vF}< 0$ on $\big(r_\Dislike(\MA),\,\widehat{r}_\Dislike(\MH)\big)$; 
    $\Delta W_{\vF}>0$ on $\big(r_\Like(\MH),\,r_\Like(\MA)\big)$.
    
    \item $\Delta W < 0$ on $\big(\widehat{r}_\Like(\MH),\,\widehat{r}_\Dislike(\MH)\big)$, if $p<\bar{p}$. 
\end{enumerate}
Here, $\bar{p}:=\frac{1}{1+(\frac{\KH}{1-\KH})^2}\in(0.5,1)$ and the reliability thresholds are given in \eqref{eq:r}--\eqref{eq:r_cond}.
\end{proposition}
\begin{proof}
    Recall that $\Delta W_v=\lam^{v}(\{\MA\})-\lam^{v}(\{\MH\})$. By \eqref{eq:r}--\eqref{eq:r_cond} and the condition on $\KH$, the thresholds of Corollary~\ref{cor:reliability_thresholds} are ordered as
    \begin{align*}
        \vT&:\ r_\Dislike(\MA)<r_\Dislike(\MH)<\widehat r_\Like(\MH)<r_\Like(\MA),\\
        \vF&:\ r_\Dislike(\MA)<\widehat r_\Dislike(\MH)<r_\Like(\MH)<r_\Like(\MA).
    \end{align*}
    (a): Outside $\big(r_\Dislike(\MA),r_\Like(\MA)\big)$, both methods lead a.s. to the same cascade. (b), (c): On each interval, $\MH$ leads a.s. to a single cascade, so $\lam^{v}(\{\MH\})\in\{0,1\}$, while $\MA$ leads to both with positive probability, so $\lam^{v}(\{\MA\})\in(0,1)$. (d): The intervals where $\Delta W_\vT<0$ and $\Delta W_\vF<0$ intersect in $\big(\widehat r_\Like(\MH),\widehat r_\Dislike(\MH)\big)$, nonempty iff $p<\bar p$ by \eqref{eq:r_cond_D}--\eqref{eq:r_cond}, and $\Delta W=r\Delta W_\vT+(1-r)\Delta W_\vF<0$ there.
\end{proof}

\begin{figure}[tbp]
\centering
\vspace{4pt}% a float at the top of a page would otherwise start above the 0.75in margin
% centre the subcaptions on the plot areas: the margins are ecc_plot.SIDE_MARGINS of code/plot
\captionsetup[sub]{justification=centering,margin={0.32in,0.04in}}
\includegraphics[width=\columnwidth]{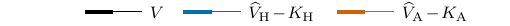}\\[2pt]
\begin{subfigure}[t]{0.49\columnwidth}
\includegraphics[width=\linewidth]{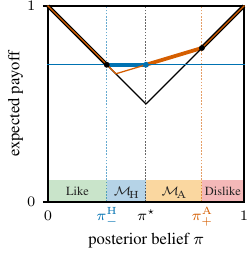}
\caption{Regime~\ref{it:both_used}, both methods are used, $\eA{1}=0.45$, $\eA{2}=0.15$.}\label{fig:sec5_cutoffs_moderate_tech}
\end{subfigure}\hfill
\begin{subfigure}[t]{0.49\columnwidth}
\includegraphics[width=\linewidth]{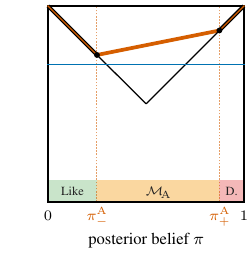}
\caption{Regime~\ref{it:automated_preferred}, $\MA$ is preferred, $\eA{1}=0.3$, $\eA{2}=0.1$.}\label{fig:sec5_cutoffs_strong_tech}
\end{subfigure}
\caption{Expected payoff of an agent with posterior $\pi$ if she does not fact-check, $V$, and if she fact-checks using method $\calM$, $\widehat V_\calM-K_\calM$, where $\widehat V_\calM$ is given in \eqref{eq:V_pi}. The highest curve (thick) is her expected payoff and the strip below shows her choice. $\pi^\star$ is where the curves of $\MH$ and $\MA$ cross. At the ends, where $\widehat V_\mathrm{A}=V$ (as $\MA$ is free), she is indifferent and does not check (see Section~\ref{subsec:individual}). This tie-break is immaterial, since no verdict of $\MA$ would change her action. \mbox{$\pi^\mathrm{H}_-:=\pi_-(\MH)$}, \mbox{$\pi^\mathrm{A}_\pm:=\pi_\pm(\MA)$}, $\KH=0.3$.}
\label{fig:sec5_cutoffs}
\end{figure}

Proposition~\ref{prop:sec4_shaded_area} shows two opposing effects of switching from $\MH$ to $\MA$. 
The first is \emph{broader checking}: since $\MA$ is free and accurate enough, agents check over a wider range of posteriors (Fig.~\ref{fig:sec5_cutoffs_strong_tech}). A truthful article of moderately low reliability, which every agent dislikes without checking under $\MH$, is now checked by some agents and escapes this wrong cascade with positive probability. So does a fake article of moderately high reliability. These are the positive bands in Figs.~\ref{fig:sec5_d_wel_rho0_p_T} and~\ref{fig:sec5_d_wel_rho0_p_F}. 

The second is \emph{misleading errors}: on the negative bands of Figs.~\ref{fig:sec5_d_wel_rho0_p_T} and~\ref{fig:sec5_d_wel_rho0_p_F}, $\MH$ leads to the correct cascade a.s., since no agent takes the wrong action without checking and $\MH$ never errs. Under $\MA$, a check wrongly verifies a fake article with probability $\eA{2}$, or wrongly falsifies a truthful one with probability $\eA{1}$, and the actions that follow can start a wrong cascade. If $p<\bar p$, i.e., truthful articles are not informative enough, the two negative bands overlap on an intermediate reliability range, as shown in Fig.~\ref{fig:sec5_d_wel_rho0_p_all}. 
% On both sides of the range, the simulation shows gains at moderately high and low reliability, where the benefit of broader checking outweighs the harm of misleading errors. At extreme reliability, $\ell^\vF_t$ starts directly in the cascade sets for both methods, so welfare does not change (zero bands in Fig.~\ref{fig:sec5_d_wel_rho0_p}). 

% ------------------------- both MH and MT used --------------------------------------
\subsection{Both Methods Are Used (Extension)}  \label{subsubsec:asym_acc_two_tools}

Regime~\ref{it:both_used} requires an extension of our model, since both methods may be used.
Denote by $\pi_{\pm}(\calM)$ the cutoffs in Proposition~\ref{prop:cutoffs} for method $\calM$.
As Fig.~\ref{fig:sec5_cutoffs_moderate_tech} shows, an agent is indifferent between the two methods at the posterior $\pi^\star:=\frac{\eA{1}-\KH}{\eA{1}-\eA{2}}$. She checks with $\MH$ if $\pi\in(\pi_-(\MH),\pi^\star]$, with $\MA$ if $\pi\in(\pi^\star,\pi_+(\MA))$, and does not check otherwise. In the following, as in Proposition~\ref{prop:sec4_shaded_area}, we certify the welfare change in some regions. The proof, along with the extended model, will appear in the journal version of this work.

\begin{proposition} \label{prop:sec5_asym_acc_two_tools_shade}
Suppose that Assumption~\ref{asmp:symmetric_news} holds and $\frac{\eA{2}}{1-\eA{1}+\eA{2}}<\KH\leq\frac{\eA{1}}{1+\eA{1}-\eA{2}}$. We have:
\begin{enumerate}[label=(\alph*)]
    \item $\Delta W_\vT=\Delta W_\vF=0$ for $r\notin\big(r_\Dislike(\MA),r^\star\big)$.
    
    \item $\Delta W_{\vT}>0$ on $\big(r_\Dislike(\MA),\,r_\Dislike(\MH)\big)$; 
    $\Delta W_{\vT}<0$ on $\big(\widehat{r}_\Like(\MH),\,r^\star\big)$.
    
    \item $\Delta W_{\vF}\leq 0$ on $\big(r_\Dislike(\MA),\, \widetilde{r}\big)$. If $p>p_\kappa$, then $\widetilde r=\widehat{r}_\Dislike(\MH)$ and the inequality is strict; otherwise $\widetilde{r}=r^\star$.
    
    \item $\Delta W < 0$ on $(\widehat{r}_\Like(\MH), \widetilde{r})$ if $p < \bar{p}$.
\end{enumerate}
Here, $\bar p$ is as in Proposition~\ref{prop:sec4_shaded_area}, the reliability thresholds are given in \eqref{eq:r}--\eqref{eq:r_cond}, $\widetilde r=\min\big(r^\star, \widehat{r}_\Dislike(\MH)\big)$, $r^\star=\frac{q\kappa}{q\kappa+(1-p)}$ and $p_\kappa=\frac{1}{1+\frac{\KH}{1-\KH}\kappa}<\bar{p}$, with $\kappa=\frac{1-\pi^\star}{\pi^\star}$.
\end{proposition}

\begin{figure}[tbp]
\centering
\vspace{4pt}% a float at the top of a page would otherwise start above the 0.75in margin
% centre the subcaptions on the plot area: the margins are _L, _R of simulation/fig3_welfare_r_p.py
\captionsetup[sub]{margin={0.34in,0.04in}}
\begin{subfigure}[t]{0.49\columnwidth}
\includegraphics[width=\linewidth]{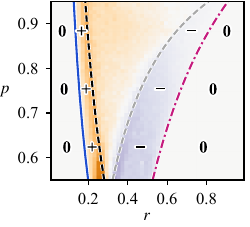}
\caption{$\Delta W_\vT$}\label{fig:sec5_d_wel_both_1_T}
\end{subfigure}\hfill
\begin{subfigure}[t]{0.49\columnwidth}
\includegraphics[width=\linewidth]{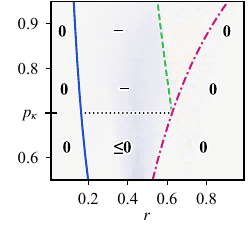}
\caption{$\Delta W_\vF$}\label{fig:sec5_d_wel_both_1_F}
\end{subfigure}\\[2pt]
\begin{subfigure}[t]{0.49\columnwidth}
\includegraphics[width=\linewidth]{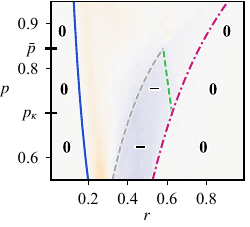}
\caption{$\Delta W$}\label{fig:sec5_d_wel_both_1_all}
\end{subfigure}\hfill
\begin{subfigure}[t]{0.49\columnwidth}
\includegraphics[width=\linewidth]{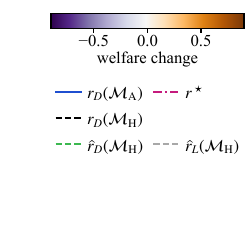}
\end{subfigure}
\caption{Welfare change in regime~\ref{it:both_used}: $\KH=0.3$, $\eA{1}=0.45$, $\eA{2}=0.15$. Lines: the thresholds \eqref{eq:r}--\eqref{eq:r_cond} and $r^\star$ bounding the regions where Proposition~\ref{prop:sec5_asym_acc_two_tools_shade} certifies the sign ($+$, $-$ strict, $\leq 0$ weak, $0$). Bold ticks mark $p_\kappa$ and $\bar p$. The horizontal dotted line in~(b) marks $p=p_\kappa$.}\label{fig:sec5_d_wel_both_1}
\end{figure}

Except for $\eA{1}=0.45$ and $\eA{2}=0.15$, the simulation setup is the same as in Section~\ref{subsec:welfare_case3}. Fig.~\ref{fig:sec5_d_wel_both_1} shows the result, with the regions certified by Proposition~\ref{prop:sec5_asym_acc_two_tools_shade} marked. The two effects of Proposition~\ref{prop:sec4_shaded_area} survive, but only one of its two positive bands does. Adding $\MA$, agents with $\pi\le\pi_-(\MH)$ still like without checking, so checking extends only to the $\Dislike$ side, from $\pi_+(\MH)$ to $\pi_+(\MA)$. Broader checking thus still helps truthful articles of moderately low reliability (Fig.~\ref{fig:sec5_d_wel_both_1_T}), but no longer fake articles of moderately high reliability, so $\Delta W_\vF$ has no certified positive band (Fig.~\ref{fig:sec5_d_wel_both_1_F}). Misleading errors remain, and the two negative bands overlap if $p<\bar p$ (Fig.~\ref{fig:sec5_d_wel_both_1_all}).

% -------------------------------------------------

\subsection{Welfare Change with an Improving Automated Method} \label{subsec:improving_tech}

Sections~\ref{subsec:welfare_case3} and~\ref{subsubsec:asym_acc_two_tools} study $\Delta W$ article by article. For a fixed $\MA$, it is a function of the reliability $r$. A platform that integrates $\MA$, however, may offer it to every user for every article, whatever the article's reliability. Yet this integration harms some articles whenever $p<\bar p$, whether only some agents use it (Proposition~\ref{prop:sec5_asym_acc_two_tools_shade}d) or every agent who checks does (Proposition~\ref{prop:sec4_shaded_area}d). As automated fact-checkers keep improving, we therefore ask when a platform should integrate $\MA$: how accurate must it be before introducing it raises welfare averaged over the circulating articles?

\subsubsection*{Simulation Setup}
Let $\overline W(\bbM):=\E_r[W(\calE,\bbM)]$, with $r$ uniformly distributed on $(0,1)$. We move $\MA$ along the accuracy relation $\eA{1}=3\eA{2}$, which contains the $\MA$ of Figs.~\ref{fig:sec5_d_wel_rho0_p} and~\ref{fig:sec5_d_wel_both_1}, and measure it by its balanced accuracy $\alpha:=1-\frac{1}{2}(\eA{1}+\eA{2})\in(1/2,1)$. The parameters are set as  $H=\mathrm{Beta}(0.5,0.5)$, $\KH=0.3$ and $p =0.55<\bar p$. We estimate $\overline W$ with $20000$ articles. At every $\alpha$, $\{\MH\}$, $\{\MA\}$ and $\{\MH,\MA\}$ face the same articles, priors and verdicts. A run ends as a cascade once the next agent's probability of liking, or of disliking, is below $10^{-5}$ under the veracity of the article. By this numerical criterion, all articles cascade within $t=1000$.

\subsubsection*{Results}
\begin{figure}[tbp]
\centering
\includegraphics[width=\linewidth]{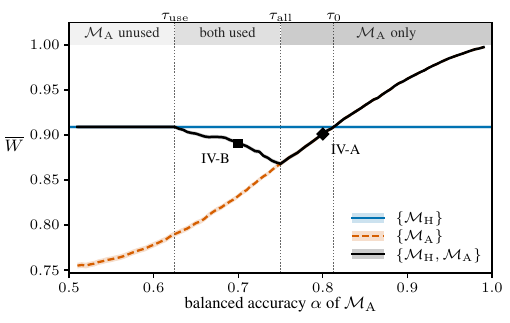}
\caption{Welfare $\overline W$ averaged over $r\sim \mathrm{Unif}(0,1)$ against the balanced accuracy $\alpha=1-\frac{1}{2}(\eA{1}+\eA{2})$ of $\MA$ along $\eA{1}=3\eA{2}$. $\KH=0.3$, $p=0.55<\bar{p}$ are fixed. Estimates are based on $20000$ articles, with $\pm1$ standard deviation shown in bands. Vertical lines: $\tau_\mathrm{use}$, below which no agent uses $\MA$; $\tau_\mathrm{all}$, above which every agent who checks does; $\tau_0$, above which $\{\MH,\MA\}$ has higher welfare than $\{\MH\}$. Top strip: regimes~\ref{it:manual_preferred}--\ref{it:automated_preferred}. Diamond, square: the $\MA$ of Sections~\ref{subsec:welfare_case3}, \ref{subsubsec:asym_acc_two_tools}.}\label{fig:sec4_welfare_vs_accuracy}
\end{figure}

In Fig.~\ref{fig:sec4_welfare_vs_accuracy}, we see that, as $\alpha$ grows, $\MA$ passes through regimes \ref{it:manual_preferred}, \ref{it:both_used} and \ref{it:automated_preferred}. Regimes~\ref{it:manual_preferred} and~\ref{it:both_used} meet at $\tau_\mathrm{use}=0.625$, where $\KH=\frac{\eA{2}}{1-\eA{1}+\eA{2}}$, and regimes~\ref{it:both_used} and~\ref{it:automated_preferred} at $\tau_\mathrm{all}=0.75$, where $\KH=\frac{\eA{1}}{1+\eA{1}-\eA{2}}$. A third threshold, $\tau_0\approx0.81$, is where the simulated $\overline W(\{\MH,\MA\})$ crosses $\overline W(\{\MH\})$. In regime~\ref{it:manual_preferred}, no agent uses $\MA$, and welfare is unchanged. In regime~\ref{it:both_used}, we observe that welfare can fall as $\MA$ improves, because more agents replace checks with $\MH$ by checks with $\MA$, whose errors mislead their successors. The dip is deepest at $\tau_\mathrm{all}$, when the last agents switch. In regime~\ref{it:automated_preferred}, welfare rises with $\alpha$, but it exceeds $\overline W(\{\MH\})$ only beyond $\tau_0$, above the $\MA$ of Figs.~\ref{fig:sec5_d_wel_rho0_p} and~\ref{fig:sec5_d_wel_both_1}.

A platform should therefore not integrate $\MA$ before it is accurate enough: until $\tau_0$, introducing it changes nothing or lowers welfare, although every agent who uses it gains privately. The loss is an informational externality~\cite{BHTW24}: an agent chooses her method for her own payoff, but her action also informs her successors, and an error of $\MA$ can start a wrong cascade among them. Even beyond $\tau_0$, where introducing $\MA$ raises welfare on average, it still lowers welfare for articles of intermediate reliability (Proposition~\ref{prop:sec4_shaded_area}d).

% Figure and its data: code/simulation/fig5_welfare_accuracy.py (its report
% prints tau_0, the dip, the two scenarios and the band term). A version where the accuracy of M_A
% depends on r was set aside on 2026-10-20: branch ECC27-accuracy-depends-on-r,
% code/simulation/MEMORY-accuracy-depends-on-r.md.

% PLAN: the appendix (threshold table + proof of Theorem thm:learning_outcomes) is
% supplementary material now; it is input after the references, see the end of this file.

%%%%%%%%%%%%%%%%%%%%%%%%%%%%%%%%%%%%%%%%%%%%%%%%%%%%%%%%%%%%%%%%%%%%%%%%%%%%%%%%

\section{CONCLUSIONS}

This paper studied how a free, error-prone automated fact-checking tool changes the probability of a correct cascade when it is offered alongside costly but accurate manual fact-checking.
The tool helps agents escape wrong cascades by bringing more of them to check, but its errors, passed on through their actions, can start wrong cascades that manual checking would have avoided. 
The second effect dominates for articles of intermediate reliability, and on average over articles the tool raises welfare only if it is accurate enough. 
A platform should therefore not offer such a tool before its accuracy exceeds a threshold.
Several questions remain for future research. First, we found the accuracy threshold above which the tool raises average welfare by simulation, but characterizing it analytically would say how accurate a tool must be before it helps. Second, a tool of given ability can trade one error rate against the other: which operating point maximizes welfare? Third, with more than two channels of information with different costs and accuracies, which should a platform make available?
\section*{APPENDIX}
\setcounter{subsection}{0}  % \section* does not reset the counter, so restart the letters at A
\renewcommand{\thesubsection}{\Alph{subsection}}  % so that \ref gives "B", not "IV-B"
\def\theHsubsection{appendix.\arabic{subsection}}  % unique hyperref anchors (else A/B clash with IV-A/IV-B and links jump there)

% From here on, number theorem-like environments as A.1, B.1, ...: the counter restarts in every appendix
% subsection and is prefixed by its letter (Theorem A.1, Lemma B.1, ...). Main-text numbering stays flat (Theorem 1, ...).
% The \theH line keeps hyperref anchors distinct from the main-text ones (else Lemma B.1 would link to Lemma 1).
\newcommand{\appendixnumbering}[1]{%
  \counterwithin{#1}{subsection}%
  \expandafter\def\csname theH#1\endcsname{\theHsubsection.\arabic{#1}}}
\appendixnumbering{theorem}
\appendixnumbering{lemma}
\appendixnumbering{proposition}
\appendixnumbering{corollary}
\appendixnumbering{definition}
\appendixnumbering{assumption}
\appendixnumbering{remark}

\subsection{Omitted Details from Section~\ref{sec:model}}\label{app:details_model}

Let $g(b):=pb+(1-p)(1-b)$, increasing in $b$ since $p>\frac12$. As in Section~\ref{subsec:individual}, $m=\stR$ and $\P_v=\P[\cdot\given m=\stR,v]$.

\subsubsection*{Beliefs~\eqref{eq:private_belief}--\eqref{eq:posterior}} A truthful article carries $m=\stR$ with probability $g(b)$ and a fake one with probability $q$, so Bayes' rule gives $\mu(b)=\frac{(1-r)q}{(1-r)q+r\,g(b)}$. Priors are independent, so given $v$ the history is independent of $b_t$ and the odds multiply: $\frac{\pi}{1-\pi}=\frac{\mu}{1-\mu}\cdot\frac{\rho_{t-1}}{1-\rho_{t-1}}$. Both $\mu$ and $\pi$ decrease in $b$.

\subsubsection*{Value of Fact-Checking~\eqref{eq:V_pi}} A verdict updates $\pi$ to $\mathrm{BU}(\pi,\vF)=\frac{\pi(1-e_2)}{\pi(1-e_2)+(1-\pi)e_1}>\pi$ or $\mathrm{BU}(\pi,\vT)=\frac{\pi e_2}{\pi e_2+(1-\pi)(1-e_1)}<\pi$, using $e_1+e_2<1$. The verdict changes her action only if $\mathrm{BU}(\pi,\vF)>\frac12\ge\mathrm{BU}(\pi,\vT)$, i.e., $\pi\in(\underline\pi,\overline\pi]$, where she follows it; elsewhere she ignores it, which gives the other two cases.

\subsubsection*{Cutoffs} In odds, $\frac{\pi}{1-\pi}=\frac{(1-r)q}{r\,g(b)}\cdot\frac{\rho}{1-\rho}$. Equating it with the odds of $\pi_\pm$ gives untruncated $b_\mp$, hence \eqref{eq:private_belief_cutoff_lo}--\eqref{eq:private_belief_cutoff_hi} after truncation. The cases $b_\pm\in\{0,1\}$ give $\ell_\Like,\ell_\Dislike,M,N$, and $\ell=1$ gives \eqref{eq:r}--\eqref{eq:r_cond}.

\subsection{Proof of Theorem~\ref{thm:learning_outcomes}}\label{app:proof_learning_outcomes}

Unless stated otherwise, we take $v=\vF$, so a.s.\ and (super)martingale statements refer to $\P_{\vF}$. Let $\F_t:=\sigma(\calH_t)=\sigma(a_1,\ldots,a_t)$, with $\F_0$ trivial, be the natural filtration of the actions. Write $\ell_t:=\ell^\vF_t$, which is $\F_t$-adapted. Recalling that $\Phi(\ell)=H(\bb_+(\ell))-H(\bb_-(\ell))$, we also define
\begin{align}
  \eta(\ell) := G_{\vF}(\ell)-G_{\vT}(\ell)
  = (1-e_1-e_2)\Phi(\ell)\ \ge 0. \label{eq:gap}
\end{align}

\begin{lemma}[a likelihood ratio is a supermartingale]\label{lem:mart}
Suppose $\P_{\vF}[a_{t+1}=a\given\F_t]=0\Rightarrow\P_{\vT}[a_{t+1}=a\given\F_t]=0$ for each action $a$. Then $(\ell_t)$ is a martingale. Without this condition, $(\ell_t)$ is a nonnegative supermartingale.
\end{lemma}
\begin{proof}
By the update rule $\ell_{t+1}=\ell_t\frac{\P_{\vT}[a_{t+1}\given\F_t]}{\P_{\vF}[a_{t+1}\given\F_t]}$, we have
\begin{align*}
  \E_{\vF}[\ell_{t+1}\given\F_t]
  &=\ell_t\sum_{a:\,\P_{\vF}[a\given\F_t]>0}\P_{\vF}[a\given\F_t]
  \frac{\P_{\vT}[a\given\F_t]}{\P_{\vF}[a\given\F_t]} \allowdisplaybreaks\\
  &=\ell_t\sum_{a:\,\P_{\vF}[a\given\F_t]>0}\P_{\vT}[a\given\F_t]. \allowdisplaybreaks
\end{align*}
The sum is $1$ under the condition and $\le 1$ otherwise.
\end{proof}

\begin{remark} \label{rem:when_martingale_fails}The condition in Lemma~\ref{lem:mart} fails when the method is perfectly sensitive, i.e., $e_2=0$. 
To see this, note that $G_{\vF}=H(\bb_+)-e_2\Phi$, and by \eqref{eq:gap}, $G_{\vT}\le G_{\vF}$, so the condition can fail only for the action $\Like$, and only if $G_{\vF}=1>G_{\vT}$. That requires $H(\bb_+(\ell))=1$ (i.e.\ $\ell\le M$), $e_2\Phi=0$ and $\Phi>0$ (i.e.\ $\ell>\ell_\Dislike$). So Lemma~\ref{lem:mart} holds at every $\ell$ \emph{except} when $e_2=0$ and $\ell\in (\ell_\Dislike, M]$. There, a fake never passes the fact-check and nobody likes without checking, so under $v=\vF$ nobody ever likes. Symmetrically, $\ell^{\vT}$ can fail to be a martingale when $e_1=0$.
\end{remark}
% \begin{remark}
% By Lemma~\ref{lem:mart} and Doob's Forward Convergence Theorem (see Williams), the nonnegative supermartingale $\ell_t$ converges to $\ell_\infty<\infty$ almost surely, and $\E_{\vF}[\ell_\infty]\leq \E_{\vF}[\ell_t] \leq \ell_{0} = 1$ for all $t\geq 1$.
% \end{remark}

% \begin{lemma}[L\'evy's extension of Borel--Cantelli]\label{lem:levy}
% For events $A_t\in\F_t$, a.s.\ $\{A_t\text{ i.o.}\}=\{\sum_t\P[A_{t+1}\given\F_t]=\infty\}$. (Williams, chapter on martingales bounded in $L^2$.)
% \end{lemma}

\begin{lemma}[a convergent limit must be stationary]\label{lem:stationary}
$\ell_\infty\in [0,\ell_\Dislike] \cup [\ell_\Like, \infty)$ a.s.
\end{lemma}
\begin{proof}
By Lemma~\ref{lem:mart}, $(\ell_t)$ is a nonnegative supermartingale, so $\ell_t\to\ell_\infty<\infty$ a.s. By \eqref{eq:gap}, $\eta(\ell)$ is continuous and strictly positive on $(\ell_\Dislike,\ell_\Like)$, since $H$ is continuous and has full support. Fix $\ell_\Dislike<\alpha<\beta<\ell_\Like$ and let $\eta_0:=\min_{[\alpha,\beta]}\eta>0$. Then for $\ell\in(\alpha,\beta)$, action $\Dislike$ has $\P_{\vF}$-probability $G_{\vF}(\ell)\ge\eta(\ell)\ge\eta_0$. After $\Dislike$, the update multiplies $\ell$ by $G_{\vT}/G_{\vF}=1-\eta/G_{\vF}\le1-\eta_0$, so $\ell_t-\ell_{t+1}\ge\eta_0\ell_t>\eta_0\alpha$. By \cite[Theorem~B.1]{SmS00}, $(\alpha,\beta)$ cannot contain any point from the support of $\ell_\infty$. (Their proof uses only the Markov property and a.s.\ convergence, so our supermartingale case is covered.) 
\end{proof}

\subsubsection*{Proof of (a)} By Lemma~\ref{lem:stationary}, $\ell_\infty\in[0,\ell_\Dislike]\cup[\ell_\Like, \infty)$ a.s. We first show that $\P_{\vF}[\ell_\infty \in \Jb_\Dislike]>0$ whenever $\ell_0\in(\ell_\Dislike,\ell_\Like)$. Suppose to the contrary that $\P_{\vF}[\ell_\infty\in\Jb_\Dislike]=0$. Then $\ell_\infty\geq \ell_\Like$ a.s., so $\E_{\vF}[\ell_\infty]\geq \ell_\Like>\ell_0$. But since $(\ell_t)$ is a supermartingale, $\E_{\vF}[\ell_\infty]\leq \ell_0$, a contradiction.

Next, suppose $\ell_0\le M$. On $(\ell_\Dislike,M]$, $G_{\vF}=1$ since $e_2=0$ (Remark~\ref{rem:when_martingale_fails}), so $a_{1}=\Dislike$ a.s.\ and $\ell_{1}=\ell_0\,G_{\vT}(\ell_0)<\ell_0$. If $\ell_t$ enters $\Jb_\Dislike$, it is frozen thereafter. Otherwise it will keep decreasing on $(\ell_\Dislike,M]$. By induction, $\ell_t<\ell_0\le M<\ell_\Like$ for all $t\ge1$, so $\ell_\infty\in\Jb_\Dislike$ a.s.\ by Lemma~\ref{lem:stationary}.

It is only left to show that $\P_{\vF}[\ell_\infty\in\Jb_\Like]>0$ if $\ell_0>M$. By Remark~\ref{rem:when_martingale_fails}, $(\ell_t)$ is not a martingale on $(\ell_\Dislike,M]$, so we stop it when it gets there. Let $\tau:=\inf\{t:\ell_t\leq M\}$ and $Y_t:=\ell_{\min\{t, \tau\}}$. For $\ell>M$, we have $\bb_+(\ell)<1$, hence $\P_{\vF}[a=\Like\given \ell] = 1-G_{\vF}(\ell)>0$. Together with $\P_{\vF}[a=\Dislike\given\ell]=G_{\vF}(\ell)\geq G_{\vT}(\ell)=\P_{\vT}[a=\Dislike\given\ell]$, the condition of Lemma~\ref{lem:mart} holds on $\{\tau>t\}\in\F_t$, hence 
% $\E_{\vF}[Y_{t+1}\given\F_t,\tau>t]=Y_t$. 
\begin{align*}
  \E_{\vF}[Y_{t+1}\given\F_t] &= \1_{\tau>t} \E_{\vF}[\ell_{t+1} \given \F_t] + \1_{\tau\leq t} \ell_\tau \\
  &= \1_{\tau>t} \ell_t + \1_{\tau\leq t} \ell_\tau = Y_t,
\end{align*}
% Moreover, $\E_{\vF}[Y_{t+1}\given\F_t,\tau\leq t]=\ell_\tau=Y_t$. Therefore $\E_{\vF}[Y_{t+1}\given\F_t]=Y_t$, 
and $(Y_t)$ is thus a martingale. Suppose $\P_{\vF}[\tau=\infty]=0$. Then $\ell_t<\ell_\Like$ a.s. for all $t\leq \tau$, otherwise it would stay in $\Jb_\Like$ forever ($\tau=\infty$) with positive probability. Thus $(Y_t)$ is bounded a.s. By the Dominated  Convergence Theorem, the mean is preserved in the limit: $\E_{\vF}[Y_\infty]=\ell_0>M$. But $Y_\infty=\ell_\tau\leq M < \ell_0$ a.s., a contradiction. Therefore, $\P_{\vF}[\tau=\infty]>0$. On $\{\tau=\infty\}$, we have $\ell_t>M>\ell_\Dislike$ for all $t$, and by Lemma~\ref{lem:stationary}, $\ell_\infty\in [\ell_\Like,\infty)$.

\subsubsection*{Proof of (b)} The proof is symmetric to (a), with the roles of $\vF$ and $\vT$ reversed and $\ell^{\vT}_t:=1/\ell_t$ for all $t$.

\subsubsection*{Proof of (c)} Let $v=\vF$ and $e_2>0$. By Remark~\ref{rem:when_martingale_fails}, $(\ell_t)$ is then a martingale, so the argument for $\ell_0>M$ in (a) applies with $M$ replaced by $(\ell_\Dislike+\ell_0)/2$, and gives $\P_{\vF}[\ell_\infty\in\Jb_\Like]>0$. Moreover, $\P_{\vF}[\ell_\infty\in\Jb_\Dislike]>0$ by the first part of (a). The case $v=\vT$, $e_1>0$ is symmetric, using $(\ell^\vT_t)$.

% \addtolength{\textheight}{-12cm}   % This command serves to balance the column lengths
                                  % on the last page of the document manually. It shortens
                                  % the textheight of the last page by a suitable amount.
                                  % This command does not take effect until the next page
                                  % so it should come on the page before the last. Make
                                  % sure that you do not shorten the textheight too much.

%%%%%%%%%%%%%%%%%%%%%%%%%%%%%%%%%%%%%%%%%%%%%%%%%%%%%%%%%%%%%%%%%%%%%%%%%%%%%%%%

%%%%%%%%%%%%%%%%%%%%%%%%%%%%%%%%%%%%%%%%%%%%%%%%%%%%%%%%%%%%%%%%%%%%%%%%%%%%%%%%

%%%%%%%%%%%%%%%%%%%%%%%%%%%%%%%%%%%%%%%%%%%%%%%%%%%%%%%%%%%%%%%%%%%%%%%%%%%%%%%%
% IEEE template boilerplate, commented out (2026-10-07): the appendix is now
% supplementary material, and funding is already acknowledged in \thanks.
% \section*{APPENDIX}
%
% Appendixes should appear before the acknowledgment.
%
% \section*{ACKNOWLEDGMENT}
%
% The preferred spelling of the word “acknowledgment” in America is without an “e” after the “g”. Avoid the stilted expression, “One of us (R. B. G.) thanks . . .”  Instead, try “R. B. G. thanks”. Put sponsor acknowledgments in the unnumbered footnote on the first page.

%%%%%%%%%%%%%%%%%%%%%%%%%%%%%%%%%%%%%%%%%%%%%%%%%%%%%%%%%%%%%%%%%%%%%%%%%%%%%%%%

% References are important to the reader; therefore, each citation must be complete and correct. If at all possible, references should be commonly available publications.

\bibliographystyle{IEEEtran}
\bibliography{bib/IEEEabrv, bib/ref.bib}

%%%%%%%%%%%%%%%%%%%%%%%%%%%%%%%%%%%%%%%%%%%%%%%%%%%%%%%%%%%%%%%%%%%%%%%%%%%%%%%%

\end{document}